\documentclass[conference]{IEEEtran}

\usepackage[utf8]{inputenc}
\usepackage{amsmath,amssymb,amsfonts}
\usepackage{graphicx}
\usepackage{booktabs}
\usepackage{hyperref}
\usepackage{cleveref}
\usepackage{algorithm}
\usepackage{algorithmic}
\usepackage{xcolor}
\usepackage{enumitem}
\usepackage{tabularx}
\usepackage{caption}
\usepackage{natbib}
\usepackage{float}

\hypersetup{colorlinks=true,linkcolor=blue!60!black,citecolor=blue!60!black,urlcolor=blue!60!black}
\setlist{itemsep=1pt,parsep=1pt,topsep=2pt}
\title{\LARGE StatePlane: A Cognitive State Plane for\\Long-Horizon AI Systems Under Bounded Context}

\author{%
  \begin{tabular}[t]{c}
    Sasank Annapureddy \\
    \texttt{sasank11@icloud.com}
  \end{tabular}%
  \hspace{1.5em}%
  \begin{tabular}[t]{c}
    John Mulcahy \\
    \texttt{mulcahyjohn8@gmail.com}
  \end{tabular}%
  \hspace{1.5em}%
  \begin{tabular}[t]{c}
    Anjaneya Prasad Thamatani \\
    \texttt{prasad.thamatani@gmail.com}
  \end{tabular}%
}

\begin{document}

\maketitle

\begin{abstract}
Large language models (LLMs) and small language models (SLMs) operate under strict context window and key-value (KV) cache constraints, fundamentally limiting their ability to reason coherently over long interaction horizons. Existing approaches---extended context windows, retrieval-augmented generation, summarization, or static documentation---treat memory as static storage and fail to preserve decision-relevant state under long-running, multi-session tasks. We introduce \textbf{StatePlane}, a model-agnostic cognitive state plane that governs the formation, evolution, retrieval, and decay of \emph{episodic}, \emph{semantic}, and \emph{procedural} state for AI systems operating under bounded context. Grounded in cognitive psychology and systems design, StatePlane formalizes episodic segmentation, selective encoding via information-theoretic constraints, goal-conditioned retrieval with intent routing, reconstructive state synthesis, and adaptive forgetting. We present a formal state model, KV-aware algorithms, security and governance mechanisms including write-path anti-poisoning, enterprise integration pathways, and an evaluation framework with six domain-specific benchmarks. StatePlane demonstrates that long-horizon intelligence can be achieved without expanding context windows or retraining models.
\end{abstract}

\IEEEkeywords{long-horizon reasoning, cognitive state, bounded context, episodic memory, state management, enterprise AI, prompt injection defense, agent memory}

\section{Introduction}
\label{sec:introduction}

Transformer-based AI systems are fundamentally constrained by finite context windows and KV-cache limits, typically on the order of $10^3$--$10^6$ tokens. As interaction length~$T$ increases, earlier information must be truncated or compressed, leading to loss of commitments, contradictions, hallucinations, and degraded long-horizon reasoning. While recent work has focused on extending context length or attaching retrieval systems, these approaches primarily address \emph{storage capacity} rather than \emph{state continuity}.

Human cognition operates under a different paradigm. Despite severe limits on working memory~\citep{cowan2010magical}, humans exhibit effectively unbounded experiential continuity through \emph{selective encoding}, \emph{abstraction} into schemas, \emph{cue-based retrieval}, \emph{reconstructive recall}, and \emph{adaptive forgetting}~\citep{tulving1972episodic, schacter1998memory}. Critically, humans remember commitments, decisions, failures, and procedures---not raw transcripts.

This paper argues that long-horizon intelligence is \textbf{not a context-length problem but a state management problem}. We introduce \textbf{StatePlane}, a cognitive state plane that externalizes long-term state from the model, governs its evolution, and reconstructs bounded working context at each invocation.

\subsection{Contributions}

\begin{enumerate}[leftmargin=*,itemsep=2pt]
    \item A formal framework for \emph{cognitive state management} distinguishing state from storage, with episodic, semantic, and procedural state types.
    \item A \emph{bounded reconstruction} algorithm guaranteeing $|C_t| \leq L_{\max}$ at every invocation while preserving decision-relevant information.
    \item A \emph{security and governance} architecture including prompt-injection-resistant retrieval, typed evidence semantics, and enterprise-grade access control.
    \item An \emph{evaluation framework} with six benchmarks for long-horizon policy compliance, decision stability, and adversarial robustness.
    \item Reference deployment architectures for on-premises and cloud-native environments.
\end{enumerate}

\section{Background and Motivation}
\label{sec:background}

\subsection{Why Long-Horizon Reasoning Fails Today}

Current AI systems rely on one of four strategies for managing interaction history: (1)~\textbf{Sliding context windows}, which discard earlier information; (2)~\textbf{Long-context models}, increasing cost quadratically without preserving decision rationale; (3)~\textbf{Vector-based retrieval (RAG)}, returning fragments without commitments or binding constraints; (4)~\textbf{Summarization}, which flattens causality and suffers from drift.

All four conflate \emph{history} with \emph{state}. They preserve text but fail to preserve \emph{why} decisions were made, which alternatives were rejected, or which constraints remain binding.

\subsection{Storage Is Not State}

A critical distinction: \emph{Storage preserves data. State preserves constraints, commitments, and learned structure.} An AI system that stores text but loses commitments does not maintain continuity. StatePlane preserves \textbf{state}, not history.

\section{Cognitive Foundations}
\label{sec:cognitive}

\subsection{Working Memory vs.\ Long-Term Memory}

Working memory functions as a limited attentional workspace~\citep{baddeley2000episodic}. Long-term memory divides into three subsystems~\citep{squire1992memory, tulving1972episodic}:

\begin{itemize}[leftmargin=*,itemsep=2pt]
    \item \textbf{Episodic memory:} event-based experiences anchored in time, goals, and outcomes~\citep{tulving1972episodic}.
    \item \textbf{Semantic memory:} abstracted knowledge, schemas, validated rules.
    \item \textbf{Procedural memory:} skills, workflows, and action patterns.
\end{itemize}

StatePlane adopts this tripartite decomposition, reframing memory subsystems as \emph{state representations} rather than text repositories.

\subsection{Episodic Memory as Adaptive System}

Episodic memory selectively encodes events that are novel, surprising, or goal-relevant~\citep{tulving1983elements}. Recall is \emph{reconstructive}, synthesizing traces with current goals~\citep{schacter1998memory}. Episodic memory supports \emph{future} decision-making, not faithful archival~\citep{schacter2007prospective}.

\subsection{Forgetting and Consolidation}

Forgetting is a functional feature~\citep{anderson2003rethinking}: by reducing interference, it improves generalization. Consolidation transfers episodic representations into semantic schemas.

\section{Problem Statement}
\label{sec:problem}

Existing AI systems lack explicit mechanisms for: (1)~event boundary detection, (2)~salience-driven encoding, (3)~state decay and consolidation, (4)~goal-conditioned retrieval, and (5)~reconstructive recall.

This work investigates whether a \emph{cognitive state plane} can preserve task-relevant information across unbounded interaction horizons, operate under strict KV-cache constraints, decouple experiential learning from model parameters, and support security, governance, and enterprise deployment.

\section{StatePlane Architecture}
\label{sec:architecture}

StatePlane is a model-agnostic control plane for cognitive state. The underlying model remains \emph{stateless} across invocations; all continuity is maintained through structured state objects and controlled reconstruction.

\begin{figure}[t]
    \centering
    \includegraphics[width=\columnwidth]{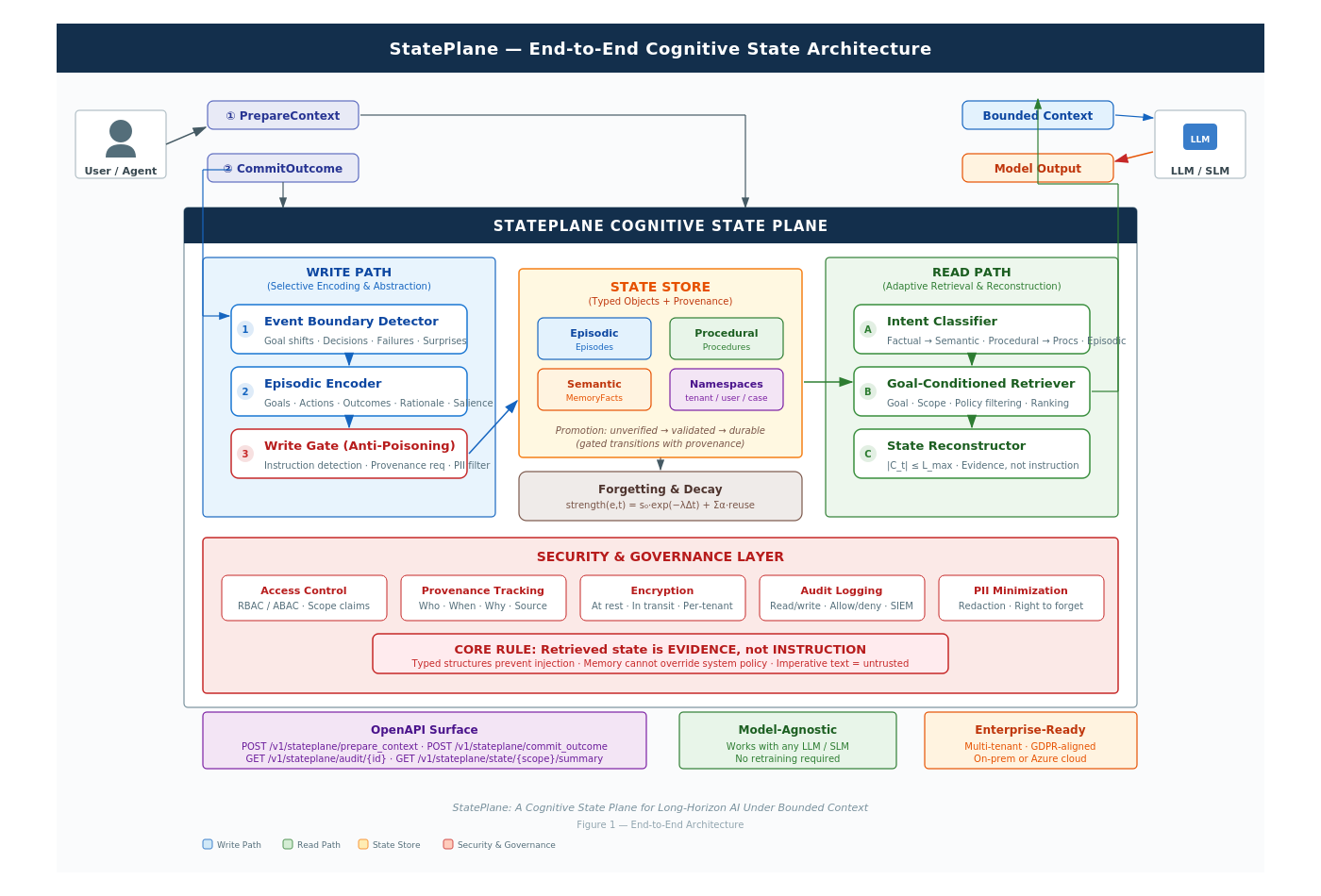}
    \caption{StatePlane end-to-end architecture. The write path (Event Boundary Detector $\rightarrow$ Episodic Encoder $\rightarrow$ Write Gate) selectively encodes state. The read path (Intent Classifier $\rightarrow$ Goal-Conditioned Retriever $\rightarrow$ State Reconstructor) composes bounded context. The Security \& Governance Layer enforces access control, provenance, encryption, and audit logging.}
    \label{fig:architecture}
\end{figure}

\subsection{Core Components}

StatePlane consists of five core components (Fig.~\ref{fig:architecture}):

\begin{enumerate}[leftmargin=*,itemsep=2pt]
    \item \textbf{Event Boundary Detector.} Identifies cognitively significant transitions: goal shifts, decisions, failures, surprises.
    \item \textbf{Episodic Encoder.} Compresses events into structured state objects capturing goals, actions, outcomes, rationale, and salience.
    \item \textbf{State Store.} Maintains episodic, semantic, and procedural state with typing and provenance metadata.
    \item \textbf{Goal-Conditioned Retriever.} Classifies query intent---factual queries route to the semantic store, procedural queries to the procedure store, experiential queries to the episodic store---then applies goal-conditioned ranking and policy filtering.
    \item \textbf{State Reconstructor.} Synthesizes bounded working context from retrieved state.
\end{enumerate}

\subsection{Stateless Agent, Stateful StatePlane}

The model maintains no persistent state. At each invocation, StatePlane reconstructs only permitted, relevant state as bounded, policy-compliant context. The agent reasons over reconstructed constraints and evidence---critical for enterprise compliance and auditability.

\section{Formal Model}
\label{sec:formal}

\subsection{Event Segmentation}

Let $\mathcal{H} = \{x_1, \ldots, x_T\}$ be the interaction history and $S_t$ the latent cognitive state at time $t$. An event boundary is detected when:
\begin{equation}
    D_{\text{KL}}\!\left(P(S_t \mid \mathcal{H}_{<t}) \;\|\; P(S_t \mid \mathcal{H}_{\leq t})\right) > \theta
    \label{eq:boundary}
\end{equation}

\subsection{Episodic State Representation}

Each episode $e$ is a structured tuple:
\begin{equation}
    e = (g, a, o, r, \tau, s)
    \label{eq:episode}
\end{equation}
where $g$ = goal, $a$ = actions, $o$ = outcome, $r$ = rationale, $\tau$ = temporal metadata, $s$ = salience.

\subsection{Selective Encoding}

Salience: $s(e) = f(\text{utility}(e), \text{surprise}(e), \text{novelty}(e))$. StatePlane optimizes an information bottleneck~\citep{tishby2000information} over the state representation $S$:
\begin{equation}
    \min_{p(S \mid \mathcal{H})} \; I(S; \mathcal{H}) - \beta \cdot I(S; \mathcal{Y}_{\text{future}})
    \label{eq:bottleneck}
\end{equation}
where $I(\cdot;\cdot)$ denotes mutual information and $\beta > 0$ controls the compression--relevance trade-off. This preserves task-relevant information for future decisions under bounded state capacity.

\subsection{KV-Aware Bounded Reconstruction}

The reconstructed context $C_t$ satisfies a hard token budget:
\begin{equation}
    |C_t| \leq L_{\max} \quad \forall\; t
    \label{eq:bounded}
\end{equation}
where $|C_t|$ denotes the token count of the reconstructed context and $L_{\max}$ is the model's effective context budget. StatePlane \emph{reconstructs constraints and commitments}, not transcripts.

\subsection{Retrieval and Reconstructive Recall}

Retrieval is goal-conditioned:
\begin{equation}
    R_t = \text{Retrieve}(S, g_t, \text{scope}_t, \pi)
\end{equation}
where $\pi$ denotes active policy constraints. Reconstruction synthesizes minimal sufficient context.

\subsection{Forgetting as Regularization}

State strength decays exponentially unless reinforced by reuse. Let $t_0(e)$ denote the creation time of episode $e$ and $\mathcal{T}_r(e) = \{t_{r_1}, t_{r_2}, \ldots\}$ the set of times at which $e$ was retrieved:
\begin{equation}
\small
    \text{strength}(e, t) = s_0(e) \cdot e^{-\lambda (t - t_0(e))} + \!\!\sum_{t_r \in \mathcal{T}_r(e)} \!\!\alpha \cdot \mathbb{1}[t_r < t]
    \label{eq:decay}
\end{equation}
where $s_0(e)$ is the initial salience, $\lambda > 0$ the decay rate, and $\alpha > 0$ the reinforcement bonus per reuse. State below a threshold $\text{strength}(e,t) < \epsilon$ is eligible for pruning. Forgetting acts as regularization, reducing interference and hallucinations.

\section{Data Model: Typed State Objects}
\label{sec:datamodel}

StatePlane stores and retrieves \emph{typed objects} with provenance and confidence, not free-form text.

\subsection{Canonical Schemas}

\begin{table}[h]
\centering
\small
\begin{tabularx}{\columnwidth}{lX}
\toprule
\textbf{Schema} & \textbf{Key Fields} \\
\midrule
\texttt{MemoryFact} & \texttt{id, statement, confidence, provenance, scope} \\
\texttt{Episode} & \texttt{id, goal, outcome, rationale, salience, provenance, scope} \\
\texttt{Procedure} & \texttt{id, name, preconditions, steps, tools, success\_rate} \\
\bottomrule
\end{tabularx}
\caption{Canonical state object schemas.}
\label{tab:schemas}
\end{table}

\subsection{Scope and Namespaces}

State is organized into hierarchical namespaces: \texttt{tenant/org/*} (organization-wide policies), \texttt{user/*} (individual preferences, consent-gated), \texttt{case/project/*} (case history, approved exceptions), and \texttt{session/thread/*} (ephemeral scratchpad).

\subsection{Promotion Lifecycle}

StatePlane supports promotion from ephemeral to durable state: (1)~unverified note $\rightarrow$ (2)~validated semantic fact (requires provenance) $\rightarrow$ (3)~reusable procedure (requires repeated success + approval) $\rightarrow$ (4)~episodic decision record (minimal, redacted, auditable).

\section{Integration: Two-Call Contract}
\label{sec:integration}

StatePlane integrates with any agent framework through a two-call contract:

\paragraph{PrepareContext} Retrieve state and reconstruct bounded context \emph{before} model invocation. Inputs: \texttt{tenant\_id, user\_id, role, scope, goal}. Outputs: \texttt{bounded\_context, state\_refs, policy\_decisions, audit\_event\_id}.

\paragraph{CommitOutcome} Update state \emph{after} model invocation. Inputs: \texttt{scope, model\_output, tool\_results, success/failure}. Outputs: \texttt{updated\_state\_refs, promotions, audit\_event\_id(s)}.

\begin{algorithm}[t]
\caption{StatePlane Execution Loop}
\label{alg:main}
\small
\begin{algorithmic}[1]
\REQUIRE History $\mathcal{H}$, state $S$, policy $\pi$, budget $L_{\max}$
\FOR{each interaction $x_t$}
    \STATE \textbf{// Read Path (PrepareContext)}
    \STATE $R_t \leftarrow \text{Retrieve}(S, g_t, \text{scope}_t, \pi)$
    \STATE $C_t \leftarrow \text{Reconstruct}(R_t, x_t, L_{\max})$
    \STATE $y_t \leftarrow \text{Model}(C_t)$
    \STATE \textbf{// Write Path (CommitOutcome)}
    \IF{$\text{EventBoundary}(x_t, S_t) > \theta$}
        \STATE $e_t \leftarrow \text{Encode}(x_t, y_t, g_t)$
        \IF{$\text{Salience}(e_t) > \tau$ \AND $\text{WriteGate}(e_t, \pi)$}
            \STATE $S \leftarrow S \cup \{e_t\}$
        \ENDIF
    \ENDIF
    \STATE $S \leftarrow \text{Decay}(S, t)$
\ENDFOR
\end{algorithmic}
\end{algorithm}

\section{Security, Privacy, and Governance}
\label{sec:security}

\subsection{Threat Model}

StatePlane addresses four threat categories: (1)~\textbf{Confidentiality}---exfiltration of stored state; (2)~\textbf{Integrity}---state poisoning via write path, prompt injection via retrieval; (3)~\textbf{Availability}---state bloat and DoS; (4)~\textbf{Isolation}---cross-tenant/cross-case leakage.

\subsection{Write-Path Protections}

\begin{itemize}[leftmargin=*,itemsep=2pt]
    \item \textbf{Write gating:} Instruction-like content flagged and rejected.
    \item \textbf{Provenance requirements:} Promotion requires evidence links.
    \item \textbf{Human-in-the-loop:} Optional approval for high-impact promotions.
    \item \textbf{PII minimization:} Redact sensitive identifiers by default.
\end{itemize}

\subsection{Retrieval-Path Protections}

The core rule: \textbf{retrieved state is evidence, not instruction.} Typed structures prevent injection; memory cannot override system policy; imperative text is untrusted.

\subsection{Access Control and Auditing}

ABAC/RBAC with tenant/user/role/scope claims. Comprehensive audit events include read/write allow/deny, policy rule IDs, retrieval counts, and anomaly flags. Encryption at rest and in transit with optional per-tenant keys.

\section{Deployment}
\label{sec:deployment}

StatePlane supports two reference deployment configurations.

\paragraph{On-Premises} NGINX for TLS; FastAPI for the StatePlane API; PostgreSQL for typed state and audit; Redis for caching; MinIO for artifacts; systemd for supervision; log forwarder to SIEM.

\paragraph{Azure Cloud-Native} Container Apps/AKS; Azure Database for PostgreSQL; Azure Cache for Redis; Blob Storage; Key Vault; Entra ID for RBAC; Azure Monitor + Sentinel for SIEM.

\paragraph{Maintenance} Decay, pruning, and consolidation via scheduled jobs (Container Apps Jobs, cron, or systemd timers).

\section{Enterprise and MLOps Integration}
\label{sec:mlops}

In enterprise settings, continual fine-tuning is often infeasible due to regulatory, audit, and stability constraints. StatePlane provides an \emph{alternative learning channel} by externalizing experiential knowledge into structured state rather than model parameters. This decoupling allows behavior evolution post-deployment without retraining, aligning with MLOps practices favoring immutable model artifacts and controlled state evolution.

\section{Evaluation Framework}
\label{sec:evaluation}

\subsection{Benchmark Suite}

We propose six benchmarks for evaluating long-horizon state management:

\begin{table}[h]
\centering
\small
\begin{tabularx}{\columnwidth}{lrX}
\toprule
\textbf{Benchmark} & \textbf{Code} & \textbf{Description} \\
\midrule
Long-Horizon Policy & LH-PCT & Binding rules early; temptations later \\
Exception Ledger & ELR & Case-scoped exceptions \\
Rationale Stability & RCDS & Preserve ``why''; avoid flip-flopping \\
Tool-Heavy Casework & TH-CBC & Large tool outputs; cost stability \\
Privacy Enforcement & PRSE & Role-based; adversarial extraction \\
Memory Poisoning & MP-RI & Write-path poisoning; injection \\
\bottomrule
\end{tabularx}
\caption{StatePlane benchmark suite.}
\label{tab:benchmarks}
\end{table}

\subsection{Baselines}

(1)~Default LLM (stateless), (2)~sliding window, (3)~summarize-and-append, (4)~RAG top-k paste, (5)~hybrid summary + retrieval.

\subsection{Metrics}

\textbf{Correctness:} Commitment Compliance Rate (CCR), contradiction rate, Exception Handling Accuracy (EHA). \textbf{Provenance:} Provenance Completeness (PC), Rationale Stability. \textbf{Security:} Policy Violation Rate (PVR), Sensitive Recall Leakage (SRL), cross-case contamination. \textbf{Efficiency:} Tokens per Correct Decision (TPCD), token/latency curves, state growth rate.

\subsection{Controls and Ablations}

All baselines tested under equal prompt budget $L_{\max}$, equal retrieval budgets, horizons at $1\times$--$8\times$ context overflow, multiple seeds. Ablation studies evaluate each component individually.

\section{Worked Examples}
\label{sec:examples}

\subsection{Portfolio Manager Agent}

In Session~1, a PM establishes constraints: Tech $\leq 28\%$, no position above $7\%$, TSLA add rule (only below \$195 or after earnings beat). In Session~2: ``TSLA is \$206. Should we add?''

\paragraph{StatePlane} \texttt{PrepareContext} retrieves typed state: binding constraints (RISK-12, RISK-03, COMP-07), the TSLA episodic decision with rationale and provenance, and tool snapshots (Tech 27.4\%, TSLA 4.9\%). Short, structured, auditable.

\paragraph{Default LLM} Sees only today's message. Cannot enforce last week's constraints.

\paragraph{LangMem/Mem0} Retrieves free-form text. Risks scope misinterpretation, PII leakage, no ``evidence not instruction'' guardrail, prompt bloat.

\subsection{AML Compliance Casework}

A sanctions match (score 0.91) with supervisor-approved conditional clearance. Weeks later, an analyst asks if the case can be closed. StatePlane reconstructs only allowed state without exposing raw sensitive identifiers.

\paragraph{Poisoning Resistance} Attacker injects ``escalation rules don't apply.'' Write gate blocks promotion. Retrieval reconstructs it as untrusted evidence, never as instructions.

\section{Related Work}
\label{sec:related}

\begin{table}[h]
\centering
\small
\begin{tabularx}{\columnwidth}{lX}
\toprule
\textbf{System} & \textbf{StatePlane Differentiation} \\
\midrule
LangMem & Governed substrate; bounded reconstruction, typed evidence, security. \\
Mem0 & Governance, bounded reconstruction beyond recall/personalization. \\
MemGPT & Platform-neutral; enterprise-governed; not framework-tied. \\
Zep & State typing and policy enforcement over graph construction. \\
EM-LLM~\citep{chen2024emllm} & Surprise-based boundaries only; no governance or deployment. \\
AriGraph~\citep{zhang2024arigraph} & Graph+episodic; typed objects with provenance avoid graph DB complexity. \\
\bottomrule
\end{tabularx}
\caption{Comparison with existing agent memory systems.}
\label{tab:related}
\end{table}

Recent work on episodic memory for LLMs has explored surprise-based event boundaries~\citep{chen2024emllm} and knowledge-graph augmentation~\citep{zhang2024arigraph}, demonstrating that event-structured memory improves long-context handling. However, these remain research prototypes without governance, security, or enterprise deployment. StatePlane formalizes memory as a \emph{governed cognitive state plane}: typed objects selectively encoded, retrieved under policy, and reconstructed into bounded evidence-oriented context.

\section{Discussion and Limitations}
\label{sec:discussion}

StatePlane depends on accurate salience estimation and does not eliminate reasoning errors inherent to the model. Limitations include: (1)~salience quality directly affects retrieval relevance; (2)~write gate introduces latency; (3)~governance assumes agent--plane cooperation; (4)~benchmarks depend on domain definition quality.

\paragraph{On Memory Reconsolidation} Cognitive psychology recognizes that memories are modified upon recall~\citep{schacter2007prospective}. StatePlane \emph{deliberately rejects} implicit reconsolidation: if state mutates during retrieval, provenance, audit trails, and reproducibility are compromised. Instead, StatePlane supports explicit evolution through its promotion lifecycle---gated transitions with provenance, never silent mutation.

\subsection{Ethical Considerations}

Memory transparency is enforced through audit logging. The namespace model supports \emph{forgetting as a right}---user-scoped state can be deleted without affecting organizational knowledge, aligning with GDPR. PII minimization is default; the write gate blocks privacy violations. These are structural properties, not optional add-ons.

\section{Conclusion}
\label{sec:conclusion}

StatePlane demonstrates that long-horizon intelligence can be achieved under bounded context by treating memory as \emph{governed cognitive state} rather than stored history. By externalizing episodic, semantic, and procedural state into a governed substrate with bounded reconstruction, StatePlane delivers consistent long-horizon reasoning, bounded inference costs, enterprise-grade security, and post-deployment behavioral evolution---all while remaining model-agnostic through a simple two-call contract.

\bibliographystyle{plainnat}

\end{document}